\documentclass[10pt,twocolumn,letterpaper]{article}

\usepackage[pagenumbers]{cvpr} % To force page numbers, e.g. for an arXiv version

\usepackage{graphicx}
\usepackage{amsmath}
\usepackage{amssymb}
\usepackage{booktabs}
\usepackage{utfsym}
\usepackage{multirow}
\usepackage{algorithm}
\usepackage{algorithmic}

\usepackage[pagebackref,breaklinks,colorlinks]{hyperref}

\usepackage[capitalize]{cleveref}
\crefname{section}{Sec.}{Secs.}
\Crefname{section}{Section}{Sections}
\Crefname{table}{Table}{Tables}
\crefname{table}{Tab.}{Tabs.}

\def\cvprPaperID{88} % *** Enter the CVPR Paper ID here
\def\confName{CVPRW}
\def\confYear{2023}

\begin{document}

\title{SAM-IQA: Can Segment Anything Boost Image Quality Assessment ?}

% \author{Lei Yu\\
% Megvii\\
% {\tt\small yulei02@megvii.com}

% \and

\author{Xinpeng Li\,$^{1}$ \thanks {Equal contribution}\quad
Ting Jiang\,$^1$ \footnotemark[1]  \quad
Haoqiang Fan\,$^1$ \quad
Shuaicheng Liu\,$^{2,1}$\thanks{Corresponding author}  \\
$^1$\,Megvii Technology\\
$^2$\, University of Electronic Science and Technology of China\\
% \tt\small \{lixinpeng\}@megvii.com, \\
% \tt\small \{jianting, wuqi02, fhq\}@megvii.com, liushuaicheng@uestc.edu.cn
}
\maketitle

%%%%%%%%% ABSTRACT
\begin{abstract}
Image Quality Assessment (IQA) is a challenging task that requires training on massive datasets to achieve accurate predictions. However, due to the lack of IQA data, deep learning-based IQA methods typically rely on pre-trained networks trained on massive datasets as feature extractors to enhance their generalization ability, such as the ResNet network trained on ImageNet. In this paper, we utilize the encoder of Segment Anything, a recently proposed segmentation model trained on a massive dataset, for high-level semantic feature extraction. Most IQA methods are limited to extracting spatial-domain features, while frequency-domain features have been shown to better represent noise and blur. Therefore, we leverage both spatial-domain and frequency-domain features by applying Fourier and standard convolutions on the extracted features, respectively. Extensive experiments are conducted to demonstrate the effectiveness of all the proposed components, and results show that our approach outperforms the state-of-the-art (SOTA) in four representative datasets, both qualitatively and quantitatively. Our experiments confirm the powerful feature extraction capabilities of Segment Anything and highlight the value of combining spatial-domain and frequency-domain features in IQA tasks. Our codes are available at \href{https://github.com/Hedlen/SAM-IQA}{https://github.com/Hedlen/SAM-IQA}.
\end{abstract}

%%%%%%%%% BODY TEXT
\section{Introduction}
% Image Quality Assessment (IQA) is a crucial task in image processing, as it enables the objective evaluation of image quality based on human preferences. Accurate IQA can benefit various image-related tasks such as super-resolution\cite{li2022d2c, yu2023dipnet, luo2021ebsr, luo2022deep, li2023ntire}, image denoising~\cite{lehtinen2018noise2noisecheng, cheng2021nbnet, wu2023realistic}, and high dynamic range imaging~\cite{liu2021adnet, liu2022ghost, perez2021ntire}. IQA approaches can be broadly classified into reference-based and reference-free methods. Reference-based methods require a reference image during evaluation, while reference-free methods do not. 
Image Quality Assessment (IQA) plays a crucial role in image processing by providing an objective evaluation of image quality based on human preferences. Accurate IQA is beneficial for various image-related tasks, including Super Resolution (SR)~\cite{li2022d2c, yu2023dipnet, luo2021ebsr, luo2022deep, li2023ntire}, Image Denoising~\cite{lehtinen2018noise2noise, cheng2021nbnet, wu2023realistic}, and High Dynamic Range (HDR) imaging~\cite{liu2021adnet, liu2022ghost, perez2021ntire}. IQA approaches can be broadly classified into two categories: reference-based and reference-free methods. Reference-based methods require a reference image as a comparison during the evaluation process. These methods measure the perceptual difference between the distorted image and the reference image to assess the quality. On the other hand, reference-free methods do not rely on a reference image and instead directly estimate the quality score based on the distorted image's features. Reference-free methods are particularly useful when a reference image is not available or practical to obtain.

% Image Quality Assessment (IQA) is a human-centric evaluation of image quality that integrates decades of human experience. However, for models to perform IQA, they require extensive training on large datasets. Existing IQA datasets often fail to provide sufficient information to support the model's judgments. As a result, current methods typically rely on pre-trained models for feature extraction as a reference for evaluation. For example, ResNet~\cite{he2016deep} pre-trained on ImageNet has been extensively utilized in various IQA approaches, consistently achieving exceptional performance. Recently, there has been a trend toward using models trained on massive datasets to improve performance. The well-known large language model (LLM)~\cite{devlin2018bert, ouyang2022training, brown2020language} has achieved revolutionary advancements in its field. Similarly, the recently proposed Segment Anything (SAM)~\cite{kirillov2023segment} is a semantic segmentation model trained on a massive dataset. Due to its vast amount of training samples and excellent generalization ability, it achieves state-of-the-art performance in comprehensive object segmentation. Its strong segmentation and feature extraction capabilities have since been widely utilized in various applications. Therefore, in this paper, we introduce the SAM for feature extraction in IQA.

IQA is a subjective evaluation process that incorporates human experience to assess image quality. However, to enable models to perform IQA, they require extensive training on large datasets. Existing IQA datasets often lack sufficient information to support accurate model judgments. Consequently, current methods rely on pre-trained models for feature extraction as a reference for evaluation. For instance, ResNet~\cite{he2016deep}, pre-trained on ImageNet, is widely used in various IQA approaches and consistently delivers outstanding performance. 

Recently, there has been a trend toward using models trained on massive datasets to improve performance. The well-known large language models (LLMs)\cite{devlin2018bert, ouyang2022training, brown2020language} have achieved revolutionary advancements in their field. Similarly, the recently proposed Segment Anything (SAM)\cite{kirillov2023segment} is a semantic segmentation model trained on an extensive dataset. With its abundant training samples and exceptional generalization ability, SAM attains state-of-the-art performance in comprehensive object segmentation. It's powerful segmentation and feature extraction capabilities have found wide application across various domains. In this paper, we introduce SAM as a feature extraction encoder in IQA. By leveraging SAM's robust training and feature extraction capabilities, we aim to enhance IQA performance. This approach allows us to leverage the strengths of SAM and exploit its potential for accurate and efficient image quality assessment.

% Current IQA methods can be classified into two types based on the type of features used: spatial-domain-based methods and frequency-domain-based methods. However, most existing methods rely on spatial-domain features, whereas frequency-domain features have been shown to better represent noise, blur, and other distortions. Previous frequency-domain-based methods often used traditional strategies like DCT. With the development of deep learning, more deep learning-based methods have been proposed for extracting frequency-domain features. The most representative method is Fourier convolution. Fourier convolution~\cite{chi2020fast} is a technique that applies Fourier transforms in neural networks. According to the spectral convolution theorem, modifying a value in the frequency domain would have a global effect on the original signal. By using Fourier convolution, we can obtain a global receptive field that captures information from the entire image, unlike spatial-domain convolutions that only consider local regions. This enables us to better capture global features and relationships between image components, which can further improve the accuracy of IQA methods.
Existing IQA methods can be categorized into spatial-domain-based and frequency-domain-based approaches, depending on the type of features utilized. While most current methods rely on spatial-domain features, it has been demonstrated that frequency-domain features can better represent various image distortions such as noise and blur. Traditional frequency-domain-based methods often employed strategies like Discrete Cosine Transform (DCT). However, with the advancements in deep learning, there has been a surge in deep learning-based methods for extracting frequency-domain features. Among these methods, Fourier convolution stands out as a prominent approach~\cite{chi2020fast}. Fourier convolution involves applying Fourier transforms within neural networks. According to the spectral convolution theorem, modifying a value in the frequency domain would have a global impact on the original signal. By incorporating Fourier convolution, we can achieve a broader receptive field that captures information from the entire image, in contrast to spatial-domain convolutions that focus on local regions. This allows us to capture global features and relationships between different components of the image, which can further improve the accuracy of IQA method.

To sum up, our main contributions include:
% \vspace{-0.4cm}
\begin{itemize}
\item We are the first to introduce the SAM encoder as a feature extractor in IQA and demonstrate its strong generalization ability in this domain.
\item We propose a method that integrates information from both frequency-domain and spatial-domain features.
\item We conducted extensive experiments on commonly used benchmarks and showed that our approach outperforms state-of-the-art methods.
\end{itemize}

\section{Related Work}
\subsection{Image Quality Assessment}
Image Quality Assessment (IQA) aims to simulate human evaluation of image quality. Traditional IQA methods can be classified into three types based on the presence or absence of a reference image and the type of reference image used: Full-Reference IQA (FR-IQA), Reduced-Reference IQA (RR-IQA), and No-Reference IQA (NR-IQA).

% FR-IQA is currently the most accurate form of IQA, and it is also the most widely used. PSNR and SSIM ~\cite{wang2004image} are the two most well-known FR-IQA methods, and we can often find their use of them in many other image-based tasks such as high dynamic range (HDR) ~\cite{liu2021adnet}, super-resolution ~\cite{luo2022deep}. Traditional FR-IQA methods generally compute the differences between the LQ and HQ images by computing the differences between pixels or some hand-designed features. Recently, deep learning-based methods have brought great improvements to FR-IQA. These methods use neural networks to extract image features, while some NN-based methods such as transformer are used to complete the IQA task ~\cite{cheon2021perceptual}. In the latest work \cite{cao2022incorporating}, a semi-supervised approach is used to make predictions with the help of unlabeled data. Since FR-IQA has strict requirements on references, it cannot be used in many scenarios, so the researchers proposed RR-IQA and NR-IQA. For example, ~\cite{zheng2021learning} propose an RR-IQA method, they first restore the LQ image and then use the restored image as the reference.
FR-IQA is currently the most accurate and widely used form of image quality assessment. PSNR and SSIM~\cite{wang2004image} are two well-known FR-IQA methods that are commonly employed in various image-based tasks, including HDR imaging~\cite{liu2021adnet} and SR~\cite{luo2022deep}. Traditional FR-IQA methods typically measure the differences between low-quality (LQ) and high-quality (HQ) images by comparing pixels or utilizing hand-designed features. In recent years, deep learning-based approaches have made significant advancements in FR-IQA. These methods utilize neural networks to extract image features, and some employ Neural Network (NN)-based models like transformers to perform the IQA task~\cite{cheon2021perceptual}. In a recent study by Cao et al.~\cite{cao2022incorporating}, a semi-supervised approach was employed, leveraging unlabeled data to make predictions. However, FR-IQA has strict requirements for reference images, which limits its applicability in many scenarios. To address this limitation, researchers have proposed RR-IQA and NR-IQA methods. For instance, Zheng et al.~\cite{zheng2021learning} introduced an RR-IQA method where they first restore the LQ image and then utilize the restored image as the reference for comparison. The advancements in FR-IQA, including the application of deep learning-based methods and the development of RR-IQA and NR-IQA approaches, have significantly improved image quality assessment, enabling more versatile usage across a wide range of applications.

 NR-IQA, in contrast, has wider application prospects, because it does not require any references during the assessment. ~\cite{kang2014convolutional} adopts the Convolution Neural Network (CNN) to the NR-IQA for the first time. Since then, many NN-based methods have been proposed ~\cite{pan2018blind,li2020norm}. ~\cite{pan2018blind} did not regress the mean opinion scores (MOS), but the quality map reduced the difficulty of the regression. Some prior knowledge is also incorporated during the evaluation~\cite{li2020norm,zhu2020metaiqa}.
% Network pre-trained on ImageNet for classification tasks can often improve the capability of feature extractions~\cite{li2020norm}. Mate-learning can further improve the pre-trained network to better cope with different types of distortion~\cite{zhu2020metaiqa}.
However, as pointed out by~\cite{sheikh2006statistical}, humans are good at comparing two images instead of looking at only one image. Therefore, no reference approaches are often less accurate than reference approaches.  
\subsection{Segment Anything.}
% Large-scale foundation models have achieved significant success in both natural language processing (NLP) and computer vision (CV) domains. These models are typically pre-trained on large datasets using specific tasks, which enables them to transfer knowledge to downstream tasks through efficient fine-tuning. SAM is a Large-scale foundation model in the field of semantic segmentation. SAM pre-trains a segmentation model that can perform interactive segmentation given prompts. By learning from one billion masks on 11 million images with 636 million parameters (ViT-H), SAM can segment all potential objects in an image without being affected by the image quality. We leverage this property to design a novel SAM-guided refinement module that enhances existing VSR methods effectively. 

% SAM has been widely adopted in various research fields and demonstrates its effectiveness and generalization. For instance, Lu et al.~\cite{lu2023can} applied SAM in video super-resolution to segment objects in the scene and mitigate the impact of large motions. Ma et al.~\cite{ma2023segment} achieved excellent segmentation results in medical imaging by utilizing SAM. Yu et al.~\cite{yu2023inpaint} utilized SAM for inpainting arbitrary objects.

Large-scale foundation models have achieved remarkable success in both natural language processing (NLP) and computer vision (CV) domains. These models are typically pre-trained on extensive datasets using specific tasks, enabling efficient fine-tuning and knowledge transfer to downstream tasks. Segment Anything (SAM) is a notable large-scale foundation model in semantic segmentation. By learning from 1 billion masks on 11 million images with 636 million parameters (ViT-H). SAM can segment all potential objects in an image without being affected by the image quality. This unique property allows us to design a novel SAM-guided refinement module that effectively enhances existing methods in various applications.

SAM has been widely adopted across different research fields, demonstrating its effectiveness and generalization capabilities. For example, Lu et al.\cite{lu2023can} have successfully applied SAM in video super-resolution tasks to segment objects in the scene and mitigate the impact of large motions. In medical imaging, Ma et al.\cite{ma2023segment} have achieved excellent segmentation results by leveraging SAM. Furthermore, SAM has been utilized by Yu et al.~\cite{yu2023inpaint} for inpainting arbitrary objects, showcasing its versatility beyond traditional segmentation tasks.

The broad adoption of SAM and its remarkable performance across diverse applications highlight its utility and effectiveness as a powerful tool in various research domains.

\begin{figure*}[htbp]
\centering
\includegraphics[width=0.95\linewidth]{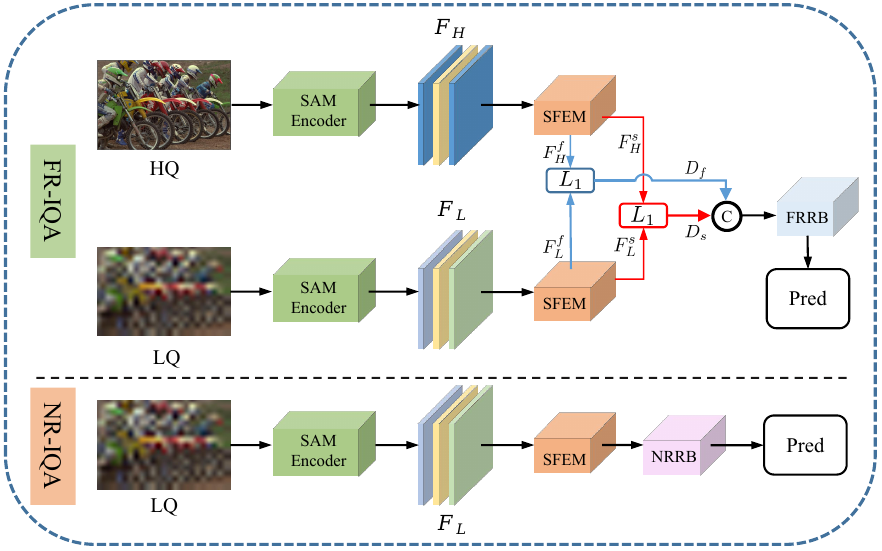}
\caption{
The pipeline of our method. The upper portion of this figure illustrates the FR-IQA process, while the lower portion describes the NR-IQA process. For FR-IQA, features are first extracted using SAM's encoder, and then spatial-domain and frequency-domain features are extracted using a spatial-frequency feature extraction module (SFEM). Distances between HQ and LQ images are then measured separately in both domains, and these distances are fed into a regression block specifically designed for full-reference IQA (FRRB) to predict the final score. In contrast, for NR-IQA, only the LQ image is utilized as input, and features are extracted using SAM's encoder. These features are further explored using SFEM, and a regression block designed for NR-IQA (NRRB) predicts the final score. Unlike FR-IQA, NR-IQA does not require any reference image for evaluation.
} \label{fig:ppl}
% \vspace{-0.5mm}
 \end{figure*}

\section{Method}

\subsection{Overall Architecture}
As shown in Figure~\ref{fig:ppl}, our proposed approach enables both full-reference~(FR) and no-reference~(NR) image quality assessment. FR-IQA is a task that evaluates the quality of an image by comparing a degraded image with a corresponding reference image. Meanwhile, NR-IQA is a task that evaluates image quality without relying on any additional information beyond the input degraded image. 

For full-reference IQA, we extract features $F_L$ and $F_H$ from the low-quality (LQ) and high-quality (HQ) images, respectively, using the encoder of SAM. Both $F_L$ and $F_H$ reside in a real space $\mathbb{R}^{256\times64\times64}$ and are fed into a spatial-frequency feature extraction module (SFEM), which is designed to extract spatial-domain and frequency-domain features and shares weights between them. The extracted features are then fed into a distance metric to obtain $D_{LH}$, which is used as input for a regression block (FRRB) specifically designed for full-reference IQA. FRRB predicts the final score.

For no-reference IQA, we first extract features from the input degraded image using the encoder of SAM. We then extract spatial-domain and frequency-domain features using SFEM and feed them into a regression block (NRRB) specifically designed for no-reference IQA. Finally, NRRB predicts the final score.

\subsection{Feature Extraction Using SAM}
Previous work has utilized networks pre-trained on ImageNet to simulate the human experience in image evaluation. In this paper, instead of using a network pre-trained on ImageNet for feature extraction, we employ the encoder of a recently proposed segmentation model trained on massive amounts of data.
\begin{equation}
\label{eq:SAM_encoder}
\begin{aligned}
F_L= \psi(I_{LQ}),\\
F_H= \psi(I_{HQ}),
\end{aligned}
\end{equation}
where, $I_{HR}$ denotes the input HQ image, $I_{LQ}$ denotes the input LQ image, $\psi$ represents the encoder of SAM, and $F_L$ and $F_H$ denote the features obtained from $I_{LQ}$ and $I_{HQ}$, respectively.

\begin{figure}[t]
\centering
\includegraphics[width=0.95\linewidth]{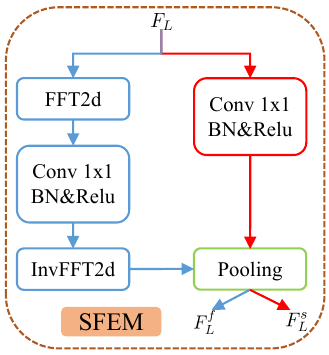}
\caption{
% The structure of SFEM. The features with different scales perform the same operations as shown in this figure. Among them, Fourier convolution is used in the left branch to extract global features, and another branch extracts local features through classical convolution. Pooling is an "AdaAvgPooling" method used to convert the results with different sizes to a uniform size.
The structure of SFEM can be visualized in the following figure. In this architecture, features of different scales undergo similar operations. Specifically, two branches are employed: one utilizes Fourier convolution on the left to extract global features, while the other employs classical convolution to extract local features. To ensure compatibility across different feature sizes, an "AdaAvgPooling" method is employed for pooling, which converts results with varying sizes into a uniform size. This allows for effective fusion and enhancement of both local and global features within the SFEM framework. 
} \label{fig:sfem}
% \vspace{-0.5mm}
\end{figure}

\subsection{Spatial-Frequency Feature Extraction Module}
After extracting semantic features using SAM, we utilize SFEM to further explore the features separately:
\begin{equation}
\label{eq:regression}
\begin{aligned}
    \{F^s_L,F^f_L\}= \{\phi_s(F_L),\phi_f(F_L)\}, \\
\{F^s_H,F^f_H\}= \{\phi_s(F_H),\phi_f(F_H)\},
\end{aligned}
\end{equation}
where, $\phi_s$ represents the feature extraction network composed of regular convolutions, while $\phi_f$ represents the feature extraction network composed of Fourier convolutions. The symbol $s$ denotes features belonging to the spatial domain, while $f$ denotes features belonging to the frequency domain. The structure diagram is shown in Figure~\ref{fig:sfem}. According to the frequency-domain convolution theorem, convolutions at each point in the frequency domain can capture the global receptive field in the spatial domain. Therefore, utilizing both frequency-domain and spatial-domain convolutions can comprehensively explore image features by combining global and local feature representations. 

\subsection{Distance metric module}
We use \(L_1\) metric to measure the feature distance between the reference image and the distorted image in the spatial and frequency domains, respectively:
% \begin{equation}
% \label{eq:regression}
% \begin{aligned}
% D_s= \xi_s(F^s_L,F^s_H), \\
% D_f= \xi_f(F^f_L,F^f_H),
% \end{aligned}
% \end{equation}
\begin{equation}
\label{eq:regression}
\begin{aligned}
D_s= L_1(F^s_L,F^s_H), \\
D_f= L_1(F^f_L,F^f_H),
\end{aligned}
\end{equation}
where, $D_s$ and $D_f$ represent the feature distance between the reference image and the distorted image in the spatial and frequency domains, respectively. This step is used only for full-reference IQA. 
\subsection{Regression}
Finally, we utilize the information obtained earlier to regress the quality score of the image, for full-reference IQA:
\begin{equation}
\label{eq:regression}
\begin{aligned}
\widetilde{y}= \Phi_f([f(D^s),g(D^f)]),
\end{aligned}
\end{equation}
 where $f$ and $g$ represent the mappings of spatial-domain and frequency-domain distances to the same distance space, respectively. $\Phi$ represents the regression network that maps the two types of distances to a quality score and $[*]$ represents the concatenation operation.

On the other hand, for NR-IQA:
\begin{equation}
\label{eq:regression}
\begin{aligned}
\widetilde{y}= \Phi_n([\eta_s(F^s_L),\eta_f(F^f_L)]),
\end{aligned}
\end{equation}
where $\eta_s$ and $\eta_f$ represent the functions that map spatial-domain and frequency-domain features to the same feature space, respectively. The operation $[*]$ denotes concatenation, and $\Phi_n$ represents the regression network that predicts the score using the features.

After obtaining the predicted score, we utilize $L_1$ loss for supervised training:
\begin{equation}
\label{eq:regression}
\begin{aligned}
loss = \lvert\widetilde{y}-y\rvert.
\end{aligned}
\end{equation}

\begin{table*}[t]
\caption{We conducted a comparison between our proposed method and other state-of-the-art approaches in two IQA tasks. The results are presented below, with the best performing results highlighted in bold and the second best results underlined. A dash (`-') indicates that the metric is not applicable.}
% \vspace{-1mm}
\centering
\setlength{\tabcolsep}{2.8mm}{
\begin{tabular}{clcccccccc}
\hline
\multirow{2}{*}{IQA Type}  & \multirow{2}{*}{Method} & \multicolumn{2}{c}{LIVE}        & \multicolumn{2}{c}{CSIQ}        & \multicolumn{2}{c}{TID2013}     & \multicolumn{2}{c}{KADID-10k}   \\
                           &                         & SRCC           & PLCC           & SRCC           & PLCC           & SRCC           & PLCC           & SRCC           & PLCC           \\ \hline
\multirow{10}{*}{FR-IQA}    & PSNR                    & 0.873          & 0.865          & 0.810          & 0.819          & 0.687          & 0.677          & 0.676          & 0.675          \\
                           & SSIM~\cite{wang2004image}                    & 0.948          & 0.937          & 0.865          & 0.852          & 0.727          & 0.777          & 0.724          & 0.717          \\
                           & VIF~\cite{sheikh2006image}                     & 0.964          & 0.960          & 0.911          & 0.913          & 0.677          & 0.771          & 0.679          & 0.687          \\
                           & PieAPP~\cite{prashnani2018pieapp}                  & 0.918          & 0.909          & 0.890          & 0.873          & 0.670          & 0.749          & 0.836          & 0.836          \\
                           & LPIPS~\cite{zhang2018unreasonable}           & 0.932          & 0.934          & 0.903          & 0.927          & 0.670          & 0.749          & 0.843     &0.839                \\
                          
                         & DISTS~\cite{ding2020image}           &0.955          & 0.955          & 0.946          & 0.946          & 0.830          &  0.855          & 0.887     & 0.886                \\
                           & IQT~\cite{cheon2021perceptual}           & 0.970          & -          & 0.943          & -          & 0.899          & -          & \underline{0.948}     &\underline{0.950}                \\
                           & ISPL-FR~\cite{cao2022incorporating}                 &\underline{0.970}          &\underline{0.978}          & \underline{0.965}          & \underline{0.968}          & 0.924          & 0.912          & 0.944          & 0.943          \\
                           & CVRKD~\cite{yin2022content}                & 0.960          & 0.965          & 0.958          & 0.965          & \underline{0.928}          & \underline{0.935}          & \textbf{0.957}         & \textbf{0.959}          \\
                           & Ours                    & \textbf{0.996} & \textbf{0.997} & \textbf{0.968} & \textbf{0.968} & \textbf{0.954} & \textbf{0.960} & 0.945 &0.947 \\ \hline

\multirow{10}{*}{NR-IQA} 
& BRISQUE~\cite{mittal2012making}         & 0.939          & 0.935          & 0.746              & 0.829              &0.604          & 0.694          & -              & -              \\
 & FRIQUEE~\cite{ghadiyaram2017perceptual}              & 0.940          & 0.944          & 0.835              & 0.874              &0.680          & 0.753          & -              & -              \\
 & BMPRI~\cite{min2018blind}              & 0.931          & 0.933          & 0.909              & 0.934              &0.928          & 0.947          & -              & -              \\
  & BIECON~\cite{kim2016fully}              & 0.961          & 0.962          & 0.815              & 0.823              &0.717          & 0.762         & -              & -              \\
   & WaDIQaM-NR~\cite{bosse2017deep}             & 0.954          & 0.963          & -              & -              &0.761          & 0.787          & -              & -              \\
& DIQaM-NR~\cite{bosse2017deep}         & 0.960          & \underline{0.972}          & -              & -              &0.835          & 0.855          & -              & -              \\
                          & IW-CNN~\cite{kim2017deep}          & \underline{0.963}          & 0.964          & 0.812          & 0.791          & 0.800          & 0.802          & -            &                \\
                          & DB-CNN~\cite{zhang2018blind}          & 0.968          & 0.971          & \textbf{0.946}          & \textbf{0.959}          & 0.816          & \underline{0.865}          & 0.501          &0.569          \\
                          & HyperIQA~\cite{su2020blindly}        & 0.962          & 0.966          & \underline{0.923}          & \underline{0.942}          &\underline{ 0.840}          & 0.858          & \textbf{0.852}              &\textbf{0.845}              \\
    & Ours        & \textbf{0.994}          & \textbf{0.995}          & 0.893          & 0.899          &\textbf{0.879}          & \textbf{0.893}          & \underline{0.815}              &\underline{0.814}              \\
                          \hline
                           
                           \hline
\end{tabular}}
\label{table:res}
\vspace{-1mm}
\end{table*}
\begin{table}[]
\caption{
Introduction to IQA datasets: The following table summarizes the characteristics of the IQA datasets used in this study. \#Ref., \#Dis., and \#Dis.Type indicate the number of reference images, distorted images, and types of distortions, respectively. }
\centering
\setlength{\tabcolsep}{1.4mm}{
\begin{tabular}{ccccc}
\hline
Dataset   & \#Ref. & \#Dis & \#Dis.Type & Score Range \\ \hline
LIVE      & 29     & 779   & 5          & {[}0,100{]} \\
CSIQ      & 30     & 866   & 6          & {[}0,1{]}   \\
TID2013   & 25     & 3000  & 24         & {[}0,9{]}   \\
KADID-10k & 81     & 10125 & 25         & {[}1,5{]}   \\ \hline
\end{tabular}}
\label{table:dataset}
\end{table}

\section{Experiments}
\subsection{Experiment Settings}
\paragraph{Datasets:} This study employs four widely-used IQA datasets, namely LIVE~\cite{sheikh2003image}, CSIQ ~\cite{larson2010most}, TID2013 ~\cite{ponomarenko2015image} and KADID-10k ~\cite{lin2019kadid}, as shown in Table~\ref{table:dataset}. LIVE and CSIQ are small-scale datasets, while TID2013 is medium-scale and KADID-10k is large-scale. The ground truth labels of TID2013 and KADIA-10k are represented by mean opinion score (MOS), while CSIQ and LIVE use differential mean opinion score (DMOS), which is inversely proportional to MOS. We adopt the common practice ~\cite{cheon2021perceptual}, ~\cite{cao2022incorporating} of randomly dividing each dataset into training, validation, and test sets with a ratio of 60\%, 20\%, and 20\%, respectively.

\paragraph{Evaluation Criterias.}
We evaluate the performance of our proposed IQA method using two commonly used metrics: Spearman's rank order correlation coefficient (SRCC) and Pearson's linear correlation coefficient (PLCC), which have been widely employed in previous IQA approaches ~\cite{cao2022incorporating,cheon2021perceptual}. SRCC assesses the accuracy of the predicted rankings of image quality, while PLCC measures the correlation between the predicted scores and the ground truth scores.
\subsection{Implementation Details}
% We employed the encoder of ViT-H, which is a large pre-trained model in SAM, to extract features for IQA. To achieve this, we adopted a two-stage process where we first extracted features using the SAM encoder and then used these saved features to train our subsequent IQA model. Note that the SAM encoder was not fine-tuned during the training process.
% During training, we augment the data by randomly applying horizontal and vertical flips to both the reference and distorted images. We set the batch size to 16, the learning rate to $2\times10^{-5}$, and trained the model using the ADAM optimizer for 500 epochs. We conducted all experiments on NVIDIA-2080ti GPUs.

We utilized the ViT-H encoder, a large pre-trained model in SAM, to extract features for IQA. Our approach involved a two-stage process: first, we extracted features using the SAM encoder, and then we employed these saved features to train our subsequent IQA model. It's important to note that the SAM encoder was not fine-tuned during the training process.

To ensure effective training, we augmented the data by applying random horizontal and vertical flips to both the reference and distorted images. The training was conducted with a batch size of 16, using the ADAM optimizer with a learning rate of $2\times10^{-5}$. The model was trained for 500 epochs on NVIDIA-2080ti GPUs.
\subsection{Comparisons with the State-of-the-art Methods}
We conduct comprehensive evaluations of our proposed method against other state-of-the-art approaches in both FR-IQA and NR-IQA settings. In the FR-IQA experiments, we compare our method with widely used metrics such as PSNR, SSIM\cite{wang2004image}, and VIF\cite{sheikh2006image}, as well as recent methods including PieAPP\cite{prashnani2018pieapp}, LPIPS\cite{zhang2018unreasonable}, IQT\cite{cheon2021perceptual}, DISTS\cite{ding2020image}, ISPL\cite{cao2022incorporating}, and CVRKD\cite{yin2022content}. The results presented in Table~\ref{table:res} demonstrate the superior performance of our method across multiple datasets, including LIVE, CSIQ, TID2013, and KADID-10k. Our method consistently outperforms other approaches in all metrics, solidifying its effectiveness and promising capabilities in IQA. 

% For NR-IQA, we compared our method with a representative and recent approaches including BRISQUE\cite{mittal2012making}, FRIQUEE\cite{ghadiyaram2017perceptual}, BMPRI\cite{min2018blind}, BIECON\cite{kim2016fully}, WaDIQaM-NR\cite{bosse2017deep}, DIQaM-NR\cite{bosse2017deep}, IW-CNN\cite{kim2017deep}, DB-CNN\cite{zhang2018blind}, and HyperIQA~\cite{su2020blindly}. Our proposed method outperformed other methods on LIVE and TID2013 datasets in all metrics, while also achieving promising results on CSIQ and KADID10k.
In the NR-IQA experiments, we compare our method with representative and recent approaches including BRISQUE\cite{mittal2012making}, FRIQUEE\cite{ghadiyaram2017perceptual}, BMPRI\cite{min2018blind}, BIECON\cite{kim2016fully}, WaDIQaM-NR\cite{bosse2017deep}, DIQaM-NR\cite{bosse2017deep}, IW-CNN\cite{kim2017deep}, DB-CNN\cite{zhang2018blind}, and HyperIQA~\cite{su2020blindly}. The evaluation was conducted on datasets such as LIVE, CSIQ, TID2013, and KADID10k. Notably, our proposed method outperformed other methods across all metrics on the LIVE and TID2013 datasets. Additionally, it achieved promising results on the CSIQ and KADID10k datasets. These findings indicate the superior performance and versatility of our method in the realm of no-reference image quality assessment. The consistent superiority over other state-of-the-art approaches showcases the potential impact and applicability of our method in various real-world scenarios.

\begin{table}[t]
\caption{We conducted a comparison between our proposed SAM encoder method and ResNet50 in the FR-IQA task. The results are shown below, with the best performing results highlighted in bold.}
\centering
\setlength{\tabcolsep}{2.2mm}{
\begin{tabular}{clcccc}
\hline
\multirow{2}{*}{Method} & \multicolumn{2}{c}{LIVE} & \multicolumn{2}{c}{TID2013}    \\
                                 & SRCC           & PLCC             & SRCC           & PLCC           \\ \hline
   ResNet50~\cite{he2016deep}                    & 0.979 & 0.982 & 0.949 & 0.958       \\
                          
                            Ours                    & \textbf{0.996} & \textbf{0.997} &  \textbf{0.954} & \textbf{0.960}  \\ \hline

% \multirow{2}{*}{NR-IQA} 
% & ResNet50~\cite{he2016deep}         & 0.962          & 0.935          & 0.746              & 0.829              &0.604          & 0.694          & -              & -              \\
%     & Ours        & \textbf{0.994}          & \textbf{0.995}          & 0.893          & 0.899          &\textbf{0.879}          & \textbf{0.893}          & \underline{0.815}              &\underline{0.814}              \\
%                           \hline
                           
%                            \hline
\end{tabular}}
\label{table:with_sam}
\end{table}

% % Please add the following required packages to your document preamble:
% % \usepackage{multirow}
% \begin{table}[]
% \caption{The SRCC of using different branch forms in SFEM. The B represents the branch that uses classical convolution to extract local features and F represents the branch that uses Fourier convolution to extract global features.}
% \centering
% \setlength{\tabcolsep}{1.6mm}{
% \begin{tabular}{cccccc}
% \hline
% IQA Type                   &  & G & \multicolumn{1}{c}{Live} & \multicolumn{1}{c}{CSIQ} & \multicolumn{1}{c}{TID} & \multicolumn{1}{c}{KADID} \\ \hline
% \multirow{3}{*}{FR-IQA}    & \usym{1F5F8}            & \usym{2717}            & 0.968                    & 0.966                    & 0.914                       & 0.962                         \\
%                            & \usym{2717}            & \usym{1F5F8}             & 0.964                    & 0.966                    & 0.937                       & 0.966                         \\
%                            & \usym{1F5F8}            & \usym{1F5F8}             & \textbf{0.979}           & \textbf{0.977}           & \textbf{0.956}              & \textbf{0.972}                \\ \hline
% \end{tabular}}

% \label{table:ablation_mlge}
% \end{table}

% \usepackage{multirow}
\begin{table}[]
\caption{The SRCC is utilized to evaluate the performance of different branch forms in SFEM based on the FR-IQA type. In this context, we denote $B$ represents the branch that uses classical convolution to extract local features and $F$ represents the branch that uses Fourier convolution to extract global features.}
\centering
\setlength{\tabcolsep}{2.2mm}{
\begin{tabular}{cccccc}
\hline
$B$ & $F$ & LIVE & CSIQ & TID2013 &LADID-10K \\ \hline
  \usym{1F5F8}            & \usym{2717}            & 0.985                    & 0.957                    & 0.912             & 0.935                                \\
                           \usym{2717}            & \usym{1F5F8}             & 0.981                    & 0.957                    & 0.935    &   0.939                                    \\
                           \usym{1F5F8}            & \usym{1F5F8}         & \textbf{0.996}           & \textbf{0.968}           & \textbf{0.954} & \textbf{0.945}                           \\ \hline
\end{tabular}}

\label{table:ablation_sfem}
\end{table}

% \usepackage{multirow}
\begin{table}[]
\caption{Comparison of results using different methods to measure the distance between distorted and reference images in FR-IQA task. Among them, \(sub.\) indicates that the acquired features are directly subtracted, while \(cos\) and \(KLD\) mean Cosine similarity and Kullback-Leibler divergence.}
\centering
\setlength{\tabcolsep}{3.2mm}{
\begin{tabular}{cccccc}
\hline
\multirow{2}{*}{Method} & \multicolumn{2}{c}{LIVE}        & \multicolumn{2}{c}{CSIQ}         \\
                         & SRCC           & PLCC           & SRCC           & PLCC           \\ \hline
                        \(Sub.\)                         & 0.971          & 0.974          & 0.950          & 0.950          \\
                         \(L_2\)                          & 0.986          & 0.988         & 0.966          & 0.962          \\
                        \(cos\)                         & 0.990          & 0.991          & 0.967          & 0.966          \\
                        \(KLD\)                         & 0.987         & 0.989          & 0.965          & 0.965          \\
                        % \(SWD\)                         & 0.967       & 0.978          & 0.966          & 0.969          \\
                         \(L_1\)                      & \textbf{0.996}          & \textbf{0.997}          & \textbf{0.968}          & \textbf{0.968}          \\ \hline
\end{tabular}}

\label{table:dis_l1}
\vspace{-3mm}
\end{table}

\subsection{Ablation Study}
\paragraph{The effectiveness of different encoders.}
We conduct comparative experiments on two encoders, including ResNet50 as our baseline and SAM's encoder. The experimental results in Table~\ref{table:with_sam} clearly demonstrate that utilizing the SAM's encoder delivers the highest performance in the FR-IQA task. This finding solidifies the superiority of our pipeline constructed with the SAM's encoder. Furthermore, the comparative analysis highlights the enhanced capabilities provided by the SAM-based approach for image quality assessment.

\paragraph{The effectiveness of Fourier Convolution in SFEM.} We conduct individual tests to evaluate the effectiveness of different branches within the SFEM framework in FR-IQA. Table~\ref{table:ablation_sfem} demonstrates that both classical convolution and Fourier convolution independently yield accurate predictions. However, when these two branches are combined, the local and global information complement each other, resulting in a further improvement in accuracy. This highlights the synergistic effect achieved by integrating both branches, allowing for more comprehensive and accurate assessments of image quality prediction.

\paragraph{The effectiveness of \(L_1\) method as distance metric.}
We assess the effectiveness of different distance measurement methods for comparing distorted and reference images, as demonstrated in Table~\ref{table:dis_l1}. We compared several commonly used methods, including \(L_2\), \(cos\), and Kullback-Leibler divergence ((KLD)). The results indicate that the \(L_1\) metric we utilized outperforms the other methods in the FR-IQA task. In scenarios involving reference frames, a simple \(L_1\) metric is already effective enough to satisfy the requirements of our method, and it has consistently delivered satisfactory results in our experiments. These findings underscore the suitability and robustness of the \(L_1\) metric for accurate image quality assessment.

% For NAR-IQA-V: we find that using Fourier convolution alone performs better than just using classical convolution. This is because distorted and reference images are not aligned, and only extracting features at the local level cannot extract enough information to determine the distortion degree of distorted images. It is always beneficial to combine local and global information on small-scale and medium-scale datasets, {\textit{i}.\textit{e}.},  LIVE, CSIQ and TID2013. However, for the large-scale dataset KADID-10k, only using Fourier convolution in MDFE is more effective than using both convolution forms at the same time. This is because the features from distorted and reference images are often completely irrelevant in the NAR-IQA-V task, which sometimes leads to some interference with the prediction. This kind of interference is especially serious when the amount of data is relatively large.
\section{Conclusion}
In conclusion, our study presents a novel approach to Image Quality Assessment (IQA) that utilizes the Segment Anything (SAM) model for feature extraction. By leveraging the vast amount of training samples and strong generalization ability of SAM, we show that our method achieves state-of-the-art performance on various IQA datasets. We also draw attention to the importance of using frequency-domain features in IQA, which has been shown to better represent various distortions present in images. Our use of Fourier convolution in neural networks allows us to capture global features and relationships between image components, leading to improved accuracy in IQA. Overall, our work demonstrates the potential of incorporating advanced deep learning techniques into IQA and highlights the value of large-scale models trained on massive datasets for this task. To showcase the effectiveness of our method, we conducted extensive experiments covering diverse datasets and evaluation metrics, providing solid evidence of its superior performance in image quality assessment.

% We have conducted comprehensive experiments to demonstrate the effectiveness of our method.

%%%%%%%%% REFERENCES
{\small
\bibliographystyle{unsrt}
\bibliography{egbib}

\begin{thebibliography}{10}

\bibitem{li2022d2c}
Youwei Li, Haibin Huang, Lanpeng Jia, Haoqiang Fan, and Shuaicheng Liu.
\newblock D2c-sr: A divergence to convergence approach for real-world image
  super-resolution.
\newblock In {\em {Proc. ECCV}}, pages 379--394, 2022.

\bibitem{yu2023dipnet}
Lei Yu, Xinpeng Li, Youwei Li, Ting Jiang, Qi~Wu, Haoqiang Fan, and Shuaicheng
  Liu.
\newblock Dipnet: Efficiency distillation and iterative pruning for image
  super-resolution.
\newblock In {\em {Proc. CVPR}}, pages 1692--1701, 2023.

\bibitem{luo2021ebsr}
Ziwei Luo, Lei Yu, Xuan Mo, Youwei Li, Lanpeng Jia, Haoqiang Fan, Jian Sun, and
  Shuaicheng Liu.
\newblock Ebsr: Feature enhanced burst super-resolution with deformable
  alignment.
\newblock In {\em {Proc. CVPR}}, pages 471--478, 2021.

\bibitem{luo2022deep}
Ziwei Luo, Haibin Huang, Lei Yu, Youwei Li, Haoqiang Fan, and Shuaicheng Liu.
\newblock Deep constrained least squares for blind image super-resolution.
\newblock In {\em {Proc. CVPR}}, pages 17642--17652, 2022.

\bibitem{li2023ntire}
Yawei Li, Yulun Zhang, Radu Timofte, Luc Van~Gool, Lei Yu, Youwei Li, Xinpeng
  Li, Ting Jiang, Qi~Wu, Mingyan Han, et~al.
\newblock Ntire 2023 challenge on efficient super-resolution: Methods and
  results.
\newblock In {\em {Proc. CVPR}}, pages 1921--1959, 2023.

\bibitem{lehtinen2018noise2noise}
Jaakko Lehtinen, Jacob Munkberg, Jon Hasselgren, Samuli Laine, Tero Karras,
  Miika Aittala, and Timo Aila.
\newblock Noise2noise: Learning image restoration without clean data.
\newblock {\em arXiv preprint arXiv:1803.04189}, 2018.

\bibitem{cheng2021nbnet}
Shen Cheng, Yuzhi Wang, Haibin Huang, Donghao Liu, Haoqiang Fan, and Shuaicheng
  Liu.
\newblock Nbnet: Noise basis learning for image denoising with subspace
  projection.
\newblock In {\em {Proc. CVPR}}, pages 4896--4906, 2021.

\bibitem{wu2023realistic}
Qi~Wu, Mingyan Han, Ting Jiang, Haoqiang Fan, Bing Zeng, and Shuaicheng Liu.
\newblock Realistic noise synthesis with diffusion models.
\newblock {\em arXiv preprint arXiv:2305.14022}, 2023.

\bibitem{liu2021adnet}
Zhen Liu, Wenjie Lin, Xinpeng Li, Qing Rao, Ting Jiang, Mingyan Han, Haoqiang
  Fan, Jian Sun, and Shuaicheng Liu.
\newblock Adnet: Attention-guided deformable convolutional network for high
  dynamic range imaging.
\newblock In {\em {Proc. CVPR}}, pages 463--470, 2021.

\bibitem{liu2022ghost}
Zhen Liu, Yinglong Wang, Bing Zeng, and Shuaicheng Liu.
\newblock Ghost-free high dynamic range imaging with context-aware transformer.
\newblock In {\em {Proc. ECCV}}, pages 344--360, 2022.

\bibitem{perez2021ntire}
Eduardo P{\'e}rez-Pellitero, Sibi Catley-Chandar, Ales Leonardis, and Radu
  Timofte.
\newblock Ntire 2021 challenge on high dynamic range imaging: Dataset, methods
  and results.
\newblock In {\em {Proc. CVPR}}, pages 691--700, 2021.

\bibitem{he2016deep}
Kaiming He, Xiangyu Zhang, Shaoqing Ren, and Jian Sun.
\newblock Deep residual learning for image recognition.
\newblock In {\em {Proc. CVPR}}, pages 770--778, 2016.

\bibitem{devlin2018bert}
Jacob Devlin, Ming-Wei Chang, Kenton Lee, and Kristina Toutanova.
\newblock Bert: Pre-training of deep bidirectional transformers for language
  understanding.
\newblock {\em arXiv preprint arXiv:1810.04805}, 2018.

\bibitem{ouyang2022training}
Long Ouyang, Jeffrey Wu, Xu~Jiang, Diogo Almeida, Carroll Wainwright, Pamela
  Mishkin, Chong Zhang, Sandhini Agarwal, Katarina Slama, Alex Ray, et~al.
\newblock Training language models to follow instructions with human feedback.
\newblock {\em {Proc. NeurIPS}}, 35:27730--27744, 2022.

\bibitem{brown2020language}
Tom Brown, Benjamin Mann, Nick Ryder, Melanie Subbiah, Jared~D Kaplan, Prafulla
  Dhariwal, Arvind Neelakantan, Pranav Shyam, Girish Sastry, Amanda Askell,
  et~al.
\newblock Language models are few-shot learners.
\newblock {\em {Proc. NeurIPS}}, 33:1877--1901, 2020.

\bibitem{kirillov2023segment}
Alexander Kirillov, Eric Mintun, Nikhila Ravi, Hanzi Mao, Chloe Rolland, Laura
  Gustafson, Tete Xiao, Spencer Whitehead, Alexander~C Berg, Wan-Yen Lo, et~al.
\newblock Segment anything.
\newblock {\em arXiv preprint arXiv:2304.02643}, 2023.

\bibitem{chi2020fast}
Lu~Chi, Borui Jiang, and Yadong Mu.
\newblock Fast fourier convolution.
\newblock {\em {Proc. NeurIPS}}, 33:4479--4488, 2020.

\bibitem{wang2004image}
Zhou Wang, Alan~C Bovik, Hamid~R Sheikh, and Eero~P Simoncelli.
\newblock Image quality assessment: from error visibility to structural
  similarity.
\newblock {\em {IEEE Trans. on Image Processing}}, 13(4):600--612, 2004.

\bibitem{cheon2021perceptual}
Manri Cheon, Sung-Jun Yoon, Byungyeon Kang, and Junwoo Lee.
\newblock Perceptual image quality assessment with transformers.
\newblock In {\em {Proc. CVPR}}, pages 433--442, 2021.

\bibitem{cao2022incorporating}
Yue Cao, Zhaolin Wan, Dongwei Ren, Zifei Yan, and Wangmeng Zuo.
\newblock Incorporating semi-supervised and positive-unlabeled learning for
  boosting full reference image quality assessment.
\newblock In {\em {Proc. CVPR}}, pages 5851--5861, 2022.

\bibitem{zheng2021learning}
Heliang Zheng, Huan Yang, Jianlong Fu, Zheng-Jun Zha, and Jiebo Luo.
\newblock Learning conditional knowledge distillation for degraded-reference
  image quality assessment.
\newblock In {\em {Proc. CVPR}}, pages 10242--10251, 2021.

\bibitem{kang2014convolutional}
Le~Kang, Peng Ye, Yi~Li, and David Doermann.
\newblock Convolutional neural networks for no-reference image quality
  assessment.
\newblock In {\em {Proc. CVPR}}, pages 1733--1740, 2014.

\bibitem{pan2018blind}
Da~Pan, Ping Shi, Ming Hou, Zefeng Ying, Sizhe Fu, and Yuan Zhang.
\newblock Blind predicting similar quality map for image quality assessment.
\newblock In {\em {Proc. CVPR}}, pages 6373--6382, 2018.

\bibitem{li2020norm}
Dingquan Li, Tingting Jiang, and Ming Jiang.
\newblock Norm-in-norm loss with faster convergence and better performance for
  image quality assessment.
\newblock In {\em Proceedings of the 28th ACM International Conference on
  Multimedia}, pages 789--797, 2020.

\bibitem{zhu2020metaiqa}
Hancheng Zhu, Leida Li, Jinjian Wu, Weisheng Dong, and Guangming Shi.
\newblock Metaiqa: Deep meta-learning for no-reference image quality
  assessment.
\newblock In {\em {Proc. CVPR}}, pages 14143--14152, 2020.

\bibitem{sheikh2006statistical}
Hamid~R Sheikh, Muhammad~F Sabir, and Alan~C Bovik.
\newblock A statistical evaluation of recent full reference image quality
  assessment algorithms.
\newblock {\em {IEEE Trans. on Image Processing}}, 15(11):3440--3451, 2006.

\bibitem{lu2023can}
Zhihe Lu, Zeyu Xiao, Jiawang Bai, Zhiwei Xiong, and Xinchao Wang.
\newblock Can sam boost video super-resolution?
\newblock {\em arXiv preprint arXiv:2305.06524}, 2023.

\bibitem{ma2023segment}
Jun Ma and Bo~Wang.
\newblock Segment anything in medical images.
\newblock {\em arXiv preprint arXiv:2304.12306}, 2023.

\bibitem{yu2023inpaint}
Tao Yu, Runseng Feng, Ruoyu Feng, Jinming Liu, Xin Jin, Wenjun Zeng, and Zhibo
  Chen.
\newblock Inpaint anything: Segment anything meets image inpainting.
\newblock {\em arXiv preprint arXiv:2304.06790}, 2023.

\bibitem{sheikh2006image}
Hamid~R Sheikh and Alan~C Bovik.
\newblock Image information and visual quality.
\newblock {\em {IEEE Trans. on Image Processing}}, 15(2):430--444, 2006.

\bibitem{prashnani2018pieapp}
Ekta Prashnani, Hong Cai, Yasamin Mostofi, and Pradeep Sen.
\newblock Pieapp: Perceptual image-error assessment through pairwise
  preference.
\newblock In {\em {Proc. CVPR}}, pages 1808--1817, 2018.

\bibitem{zhang2018unreasonable}
Richard Zhang, Phillip Isola, Alexei~A Efros, Eli Shechtman, and Oliver Wang.
\newblock The unreasonable effectiveness of deep features as a perceptual
  metric.
\newblock In {\em {Proc. CVPR}}, pages 586--595, 2018.

\bibitem{ding2020image}
Keyan Ding, Kede Ma, Shiqi Wang, and Eero~P Simoncelli.
\newblock Image quality assessment: Unifying structure and texture similarity.
\newblock {\em {IEEE Trans. on Pattern Analysis and Machine Intelligence}},
  2020.

\bibitem{yin2022content}
Guanghao Yin, Wei Wang, Zehuan Yuan, Chuchu Han, Wei Ji, Shouqian Sun, and
  Changhu Wang.
\newblock Content-variant reference image quality assessment via knowledge
  distillation.
\newblock {\em arXiv preprint arXiv:2202.13123}, 2022.

\bibitem{mittal2012making}
Anish Mittal, Rajiv Soundararajan, and Alan~C Bovik.
\newblock Making a “completely blind” image quality analyzer.
\newblock {\em IEEE Signal processing letters}, 20(3):209--212, 2012.

\bibitem{ghadiyaram2017perceptual}
Deepti Ghadiyaram and Alan~C Bovik.
\newblock Perceptual quality prediction on authentically distorted images using
  a bag of features approach.
\newblock {\em Journal of vision}, 17(1):32--32, 2017.

\bibitem{min2018blind}
Xiongkuo Min, Guangtao Zhai, Ke~Gu, Yutao Liu, and Xiaokang Yang.
\newblock Blind image quality estimation via distortion aggravation.
\newblock {\em IEEE Trans. on Broadcasting}, 64(2):508--517, 2018.

\bibitem{kim2016fully}
Jongyoo Kim and Sanghoon Lee.
\newblock Fully deep blind image quality predictor.
\newblock {\em IEEE Journal of selected topics in signal processing},
  11(1):206--220, 2016.

\bibitem{bosse2017deep}
Sebastian Bosse, Dominique Maniry, Klaus-Robert M{\"u}ller, Thomas Wiegand, and
  Wojciech Samek.
\newblock Deep neural networks for no-reference and full-reference image
  quality assessment.
\newblock {\em {IEEE Trans. on Image Processing}}, 27(1):206--219, 2017.

\bibitem{kim2017deep}
Jongyoo Kim, Hui Zeng, Deepti Ghadiyaram, Sanghoon Lee, Lei Zhang, and Alan~C
  Bovik.
\newblock Deep convolutional neural models for picture-quality prediction:
  Challenges and solutions to data-driven image quality assessment.
\newblock {\em IEEE Signal processing magazine}, 34(6):130--141, 2017.

\bibitem{zhang2018blind}
Weixia Zhang, Kede Ma, Jia Yan, Dexiang Deng, and Zhou Wang.
\newblock Blind image quality assessment using a deep bilinear convolutional
  neural network.
\newblock {\em IEEE Trans. on Circuits and Systems for Video Technology},
  30(1):36--47, 2018.

\bibitem{su2020blindly}
Shaolin Su, Qingsen Yan, Yu~Zhu, Cheng Zhang, Xin Ge, Jinqiu Sun, and Yanning
  Zhang.
\newblock Blindly assess image quality in the wild guided by a self-adaptive
  hyper network.
\newblock In {\em {Proc. CVPR}}, pages 3667--3676, 2020.

\bibitem{sheikh2003image}
Hamid~R Sheikh.
\newblock Image and video quality assessment research at live.
\newblock {\em http://live. ece. utexas. edu/research/quality}, 2003.

\bibitem{larson2010most}
Eric~Cooper Larson and Damon~Michael Chandler.
\newblock Most apparent distortion: full-reference image quality assessment and
  the role of strategy.
\newblock {\em Journal of electronic imaging}, 19(1):011006, 2010.

\bibitem{ponomarenko2015image}
Nikolay Ponomarenko, Lina Jin, Oleg Ieremeiev, Vladimir Lukin, Karen
  Egiazarian, Jaakko Astola, Benoit Vozel, Kacem Chehdi, Marco Carli, Federica
  Battisti, et~al.
\newblock Image database tid2013: Peculiarities, results and perspectives.
\newblock {\em Signal processing: Image communication}, 30:57--77, 2015.

\bibitem{lin2019kadid}
Hanhe Lin, Vlad Hosu, and Dietmar Saupe.
\newblock Kadid-10k: A large-scale artificially distorted iqa database.
\newblock In {\em 2019 Eleventh International Conference on Quality of
  Multimedia Experience (QoMEX)}, pages 1--3, 2019.

\end{thebibliography}
}

\end{document}